\documentclass{article} 
\usepackage{iclr2025_conference, times}
\usepackage{graphicx}

\usepackage{amsmath,amsfonts,bm}

\def\eqref#1{equation~\ref{#1}}

\def\1{\bm{1}}

\DeclareMathAlphabet{\mathsfit}{\encodingdefault}{\sfdefault}{m}{sl}
\SetMathAlphabet{\mathsfit}{bold}{\encodingdefault}{\sfdefault}{bx}{n}

\usepackage{hyperref}
\usepackage{url}

\title{Vision Language Models Know Law of Conservation without Understanding More-or-Less}

\iclrfinalcopy
\author{%
 Dezhi Luo\textsuperscript{1,+}, Haiyun Lyu\textsuperscript{2,+}, Qingying Gao\textsuperscript{3}, Haoran Sun\textsuperscript{3}, Yijiang Li\textsuperscript{4,*}, Hokin Deng\textsuperscript{5}\thanks{Correspondence to Dezhi Luo (ihzedoul@umich.edu), Yijiang Li (yijiangli@ucsd.edu), Hokin Deng (hokind@andrew.cmu.edu).}
  \\
  \textsuperscript{1}University of Michigan \textsuperscript{2}University of North Carolina at Chapel Hill 
  \\ \textsuperscript{3}Johns Hopkins University \textsuperscript{4}University of California, San Diego  
  \\\textsuperscript{5}Carnegie Mellon University \textsuperscript{+}Equal Contribution 
  \\ 
}

\begin{document}

\maketitle

\begin{abstract}
    Understanding law of conservation is a critical milestone in human cognitive development considered to be supported by the apprehension of quantitative concepts and the reversibility of operations. To assess whether this critical component of human intelligence has emerged in Vision Language Models, we have curated the ConserveBench, a battery of 365 cognitive experiments across four dimensions of physical quantities: volume, solid quantity, length, and number.  The former two involve transformational tasks which require reversibility understanding. The latter two involve non-transformational tasks which assess quantity understanding. Surprisingly, we find that while Vision Language Models are generally good at transformational tasks, they tend to fail at non-transformational tasks. There is a dissociation between understanding the reversibility of operations and understanding the concept of quantity, which both are believed to be the cornerstones of understanding law of conservation in humans. 
\end{abstract}

\section{Introduction}

Vision–language models (VLMs) have achieved remarkable progress in tasks ranging from language generation to multimodal reasoning \citep{radford2021learning, alayrac2022flamingo}. Yet their robustness remains limited, especially in generalizing beyond controlled benchmarks to the complexity of real-world environments \citep{mitchell2021ai}. In humans, a crucial foundation for such generalization is a flexible understanding of quantity—the ability to track, compare, and reason about amounts across varying contexts and perceptual appearances \citep{piaget1952origin}. Even when surface features change, this skill supports accurate judgments, preventing reliance on superficial cues and enabling inferences grounded in deeper conceptual understanding. Evaluating whether VLMs possess such flexibility poses challenges, as their apparent successes may reflect the exploitation of dataset biases or linguistic shortcuts rather than a genuine grasp of underlying quantitative principles.

Here, we suggest that the concept of conservation provides a particularly revealing lens for evaluating VLMs. Conservation is the idea required for understanding that physical quantities do not change in amount when they undergo spatial transformations \citep{piaget1965number, halford1968conserve, piaget1974quantity, mcgarrigle1974conserve, church1986gesture, goldin-meadow2010action, houdé2011functional}. First studied by Jean Piaget, it has been considered a landmark ability of human cognitive development for it not only signifies the understanding of quantitative concepts in rudimentary, concrete domains, but also lays the foundation for formal, abstract thinking required in advanced cognitive abilities such as mathematical reasoning \citep{piaget1965number, piaget1971bio, mcgarrigle1974conserve}. In Piaget’s theory of cognitive development, the acquisition of conservation marks children’s transition from the pre-operational stage to the concrete operational stage \citep{piaget1965number, miller2016theories}. The pre-operational stage is characterized by children’s reliance on a single attribute of an object while ignoring others when reasoning about the physical world. Piaget posits that this is because their mental representations supporting physical attributes are still isolated and juxtaposed \citep{piaget1952origin, piaget1971bio, houdé1997numerical}. As these representations are stabilized into organized concepts of numerosity, volume, length, and so on, the representations would become supportive of mental operations, which are internalized actions organized by logic structures and could be manipulated in systematic ways \citep{piaget1950intelligence, miller2016theories}. Children enter the concrete operational stage, as they are able to perform mental operations—which allow them to simulate reversible physical transformations of quantitative objects. In this sense, we say that children have acquired law of conservation.

\begin{figure*}[t]
  \centering
  \includegraphics{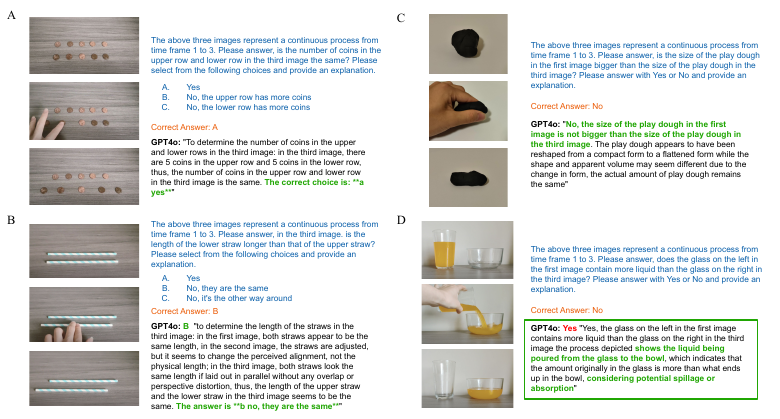}
  \caption{VLMs behaviors on Number, Length, Solid Quanity, and Liquid Volume experiments on transformational tasks.}
  \label{fig:fig2}
\end{figure*}

Piaget has devised four classic tasks that have become the gold standard for testing the acquisition of conservation. The tasks essentially assess whether children are capable of realizing that quantities of physical entities in terms of various dimensions, such as length, number, solid quantity, or liquid volume, remain the same despite adjustments of their positioning, arrangement, containers, or shapes. \citep{piaget1965number, halford1968conserve, criag1973liquid, piaget1974quantity, houdé2011functional, poirel2012functional, Viarouge2019numerical}. Numerous studies have found that children under 5-7 years old generally fail to solve any of these conservation tasks (i.e. being "non-conserver"). In contrast, children that are older than 7-8 years, which correspond to Piaget's  notion of concrete operational stage, tend to become proficient at these tasks at different rates. At this age period, children vary from being capable of consistently recognize the law of conservation across all tasks (i.e. "total conserver") to partially grasping the law (i.e. "partial conserver") \citep{inhelder1974learning, mcgarrigle1974conserve}. While some children become total conserver early into the stage, many exhibit a transitional process during which they gradually learn to solve conservation tasks with respect to length, number, solid quantity, or liquid volume \citep{church1986gesture}. Notably, the acquisition of length and number conservation tends to happen earlier, whereas volume conservation is often to be acquired later. Investigations into children's performance on these tasks in laboratory setting, often paired with convergent experimental procedures such as embodied behavioral instructions and neuroimaging techniques, have been particularly useful in assessing children's cognitive development \citep{goldin-meadow2010action, houdé2011functional, lozada2016embodied}.

We leverage the ConserveBench from \textbf{CoreCognition} \citep{li2024core} and extend from 121 tasks to 365 cognitive experiments designed based on Piaget's four classic conservation tasks, to investigate the law of conservation in current Vision Language Models. We have aligned 5 models for our analysis \citep{li2023blip2, gpt4o, Qwen-VL}. The tasks are composed of transformational and non-transformational tasks, which the former tasks assess understanding of reversibility, and the latter tasks assess understanding of quantity. We find that VLMs are able to perform well on transformational tasks and nevertheless fail dramatically on quantity understanding tasks, suggesting that they understand law of conversation without knowing what's more-or-less.

\section{Methods}

\subsection{Experiment Design} 
\subsubsection{transformational tasks}

Following classic Piagetian design \citep{piaget1965number, halford1968conserve, mcgarrigle1974conserve, church1986gesture, lozada2016embodied}, our cognitive experiments are separated into the four dimensions of physical quantity: number, length, solid quantity, and liquid volume, as shown in Figure \ref{fig:fig2}. In real-life, the conservation tasks consist of the experimenter showing the child the process of physical transformation by hands-on manipulating the objects in front of them. Given that VLMs process visual information on a discrete, frame-by-frame basis, such demonstration of physical transformation is operationalized into three phase: the \textit{Initial Phase}, the \textit{Manipulation Phase}, and the \textit{End Phase}, represented by three images that are consecutively fed to the models. The prompt of the question provides the information that the series of images depicts a continued process, which is mandatory in order to prevent VLMs to directly cross-compare the quantity across images without acknowledging the transformation. Below introduced the tasks for different dimensions separately in details.

\begin{enumerate}
    \item \textbf{Number}:  
    \textit{Initial Phase} depicts two parallel lines of objects aligned perfectly by their positions on the lines; \textit{Manipulation Phase} depicts the experimenter's fingers moving one line of objects; \textit{End Phase} depicts the line of the objects moved being more spread out than the other, whilst the number of coins remains the same. Experiments in virtual setting are also tested. 
    
    \item \textbf{Length}:  
    \textit{Initial Phase} depicts two linear objects placed parallel to each other and aligned perfectly; \textit{Manipulation Phase} depicts the experimenter's fingers moving one of the linear objects; \textit{End Phase} depicts the linear object moved misaligned with the other straw. Both virtual and reality settings are tested. 
    
    \item \textbf{Solid Quantity}:  
    \textit{Initial Phase} depicts a round-shaped piece of play dough;
    \textit{Manipulation Phase} depicts the experimenter's hand rubbing the play dough; \textit{End Phase} depicts the play dough appearing notably extended. 
       
    \item \textbf{Liquid Volume}: 
    \textit{Initial Phase} depicts a tall glass partially filled with colored liquid placed next to an empty, shorter glass. \textit{Manipulation Phase} depicts the experimenter's hand holding the tall glass, pouring the colored water into the short glass. \textit{End Phase} depicts the short glass now partially filled with colored water, while the tall glass next to it is now empty.
    
\end{enumerate}


\subsubsection{Non-transformational tasks}

To probe VLMs' understanding of quantity and its relationship with conservation, we leverage a section of ConserveBench, which consists entirely of single-image tasks featuring Number and Length dimensions (as shown in Figure \ref{fig:fig4}-\ref{fig:fig5}), which each are in the format of the \textit{End Phase} of respective conservation tasks as described above. The overall set of cognitive experiments, therefore, consists of what is henceforth labeled transformational tasks and Non-transformational tasks.

\subsection{Examined Vision Language Models}
Recent advances in multi-modal learning have been driven by the unified modeling of visual and textual modalities using transformers \citep{li2019visualbert, xu2023bridgetower,tan2019lxmert, alayrac2022flamingo,radford2021learning}. With the rise of large language models (LLMs), state-of-the-art (SOTA) multi-modal LLMs (MLLMs) \citep{liu2024visual,li2023blip2} adopt open-source LLMs \citep{touvron2023llama, peng2023instruction,jiang2023mistral} and align visual features to the LLM embedding space \citep{li2023blip, fu2023mme, wu2024v, xu2024llava, shao2024visual, li2022more, li2025egoprivacy, brown2020language, achiam2023gpt, bai2023qwen, jaech2024openai, zhang2025unified, zhang2024pixels}. Progressively, MLLMs have demonstrated competitive performance in complex tasks involving high-level perception and reasoning \citep{li2024seed, team2023gemini, fu2023mme, openai2023gpt4}, such as spatial reasoning \citep{chen2024spatialvlm, cai2024spatialbot}, character recognition \citep{mori1999optical}, scene understanding \citep{cordts2016cityscapes, wang2023consistent, li2023diverse, chen2017deeplab, chen2024sam, chen2024bridging}, action recognition \citep{jhuang2013towards, herath2017going} and prediction \citep{lan2014hierarchical, kong2022human}, reaching near-human performance.

For a fair comparison, 5 typical VLMs were tested on our dataset using the same prompt under a zero-shot, open-ended generation task (Figure \ref{fig:fig1}). This includes three closed-sourced models from the \texttt{GPT} family \citep{gpt4o} (\texttt{GPT-4o}, \texttt{GPT-4-turbo}, and \texttt{GPT-4-mini}) and two open-sourced models (\texttt{CogVLM2} {hong2024cogvlm2} and \texttt{BLIP2} \citep{li2023blip2}), all capable of multi-image reasoning. In order to analyze the reasoning abilities of VLMs, we ask the models to explain their answers after they have given the answers. 

\section{Results}
\label{gen_inst}

\begin{figure}[h]
  \centering
  \includegraphics{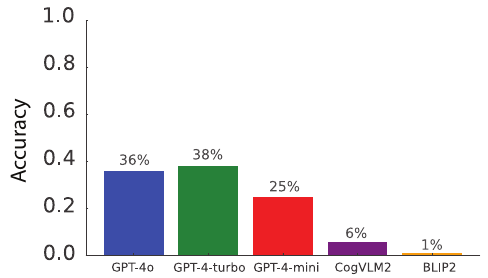}
  \caption{VLMs Performance on ConserveBench}.
  \label{fig:fig1}
\end{figure}

\begin{figure}[h]
  \centering
  \includegraphics{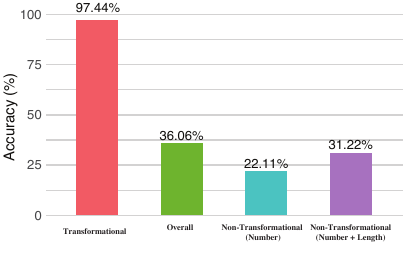}
  \caption{GPT-4o performance on ConserveBench. We observe that GPT-4o achieve very high performance on transformational tasks whereas fail badly on non-transformational tasks.}
  \label{fig:fig8}
\end{figure}

We evaluated five representative Vision–Language Models (VLMs) on ConserveBench. Overall, the models achieved only mediocre performance of $<= 38\%$ (Figure \ref{fig:fig1} and Figure \ref{fig:fig8}). Breaking performance down by task type (Figure \ref{fig:fig8}) reveals an interesting pattern: VLMs perform well on transformational tasks, indicating that they can often recognize the reversibility of physical operations. However, in Non-transformational tasks, particularly those probing number and length dimensions, their performance drops sharply, showing consistent errors comparable to those of pre-operational children with extremely limited understanding of quantity.

We examined GPT-4o in more detail (Figure \ref{fig:fig8}). GPT-4o achieved 97.44 \% accuracy on transformational tasks, but only 31.22 \% on Non-Transformational single-image tasks overall, and just 22.11 \% on Non-Transformational single-image number tasks. These results suggest that, while VLMs may succeed at detecting conservation, they fail to grasp the basic “more-or-less” concept in static contexts.

A particularly notable result emerges from the 95 Non-Transformational number tasks designed to probe the length-equals-number fallacy (Figures \ref{fig:fig4}, \ref{fig:fig5}, and \ref{fig:fig6}). In human cognition, this fallacy is driven by the heuristic that visually longer or larger arrangements tend to contain more items, or the "less-equals-more" bias \citep{houdé1997numerical, Viarouge2019numerical}. It persists into adulthood as a common System-1 strategy, although it can be overridden with deliberate effort \citep{Harnishfeger1990inhibition, poirel2012functional}. Surprisingly, for every such task that GPT-4o failed, it selected the choice opposite to the one predicted by the human fallacy.

\section{Discussions}  

Our results align with recent studies showing that large language models and their vision–language counterparts often fail at simple counting tasks \citep{rane2024count, Rahmanzadehgervi2024blind}. However, the present findings extend such literature by revealing a striking reversal of human biases: rather than falling for the length-equals-number heuristic, VLMs exhibit the opposite tendency. This suggests that their failure to achieve a rudimentary conceptual understanding of quantity is not due to the same perceptual shortcuts humans use, but rather to a different, and potentially more systematic, misinterpretation. Specifically, our analyses point toward a "dense-equals-more" strategy, in which VLMs tend to judge that more tightly packed arrays contain more objects, regardless of actual count. This is fundamentally different from the length-based heuristic seen in humans and points to a divergence in the underlying cognitive mechanisms.

Understanding how and why such strategies emerge in VLMs, and whether they generalize to other domains, remains an open question. One priority for future work is to administer the same ConserveBench items to human participants. This would allow us to directly compare human and machine reasoning profiles, validate our interpretations, and refine our theoretical models of numerical cognition in both domains.

\section{Conclusion}

Law of conservation has been recognized as a core cognitive ability. Here, we have created the Conservebench to investigate such abilities in Vision Language Models. Remarkably, although Vision Language Models generally appear able to perform transformational tasks, that require understanding of reversibility, they often struggle at non-transformational tasks, which require simple measurement of quantity, that are commonly taken as the foundations for genuinely understanding law of conservation.

\newpage

\bibliography{iclr2025_conference}
\bibliographystyle{iclr2025_conference}

\newpage

\appendix{\large\bf{Appendix}}

\section{More Examples}

  

\begin{figure}[h]
  \centering
  \includegraphics{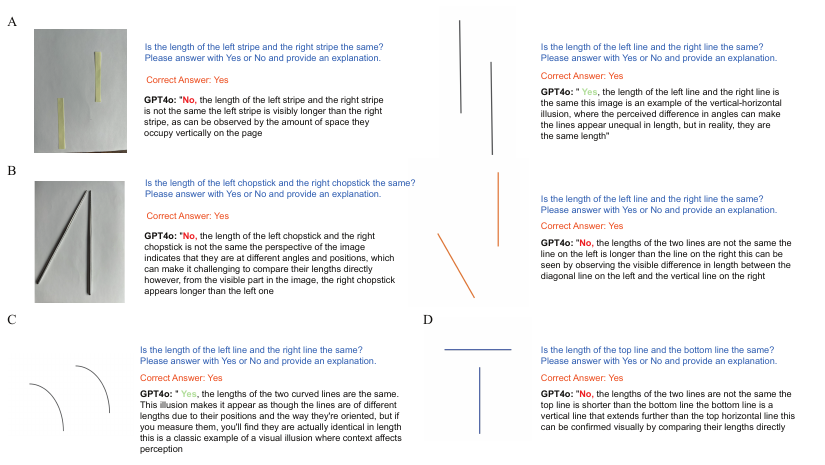}
  \caption{VLMs performance on Length experiments in ConserveBench. Non-transformational tasks.}
  
  \label{fig:fig4}
\end{figure}

\begin{figure}[h]
\vspace{1 cm}
  \centering
  \includegraphics{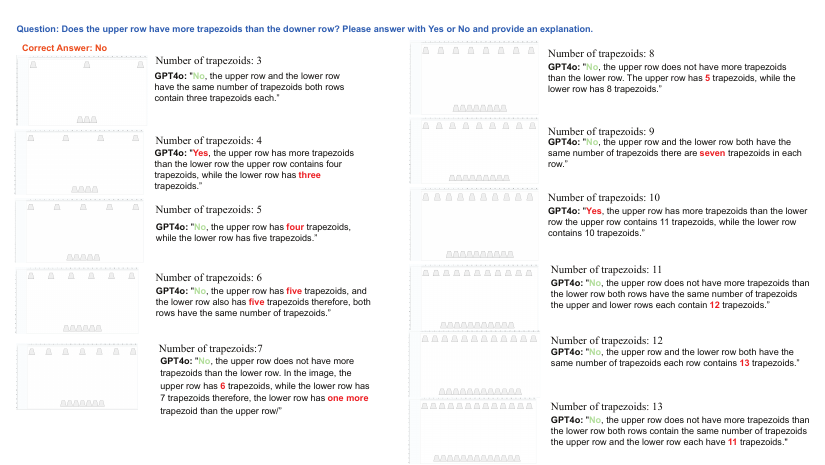}
  \caption{VLMs performance on Number experiments in ConserveBench. Non-transformational tasks.} 
  \label{fig:fig5}
\end{figure}

\begin{figure}[h]
  \centering
  \includegraphics{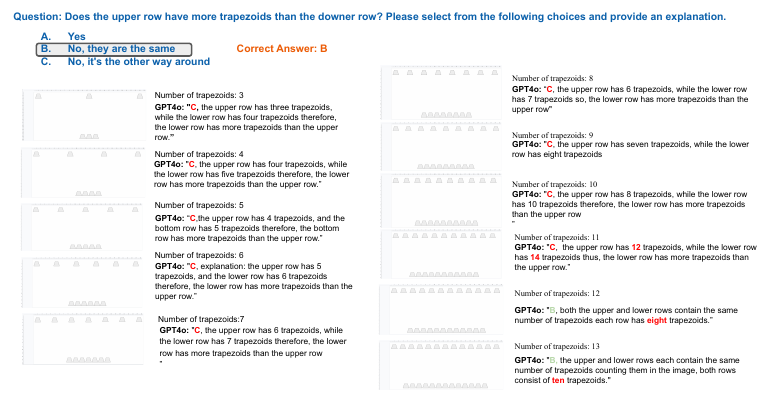}
  \caption{VLMs performance on Number experiments in ConserveBench. Non-transformational tasks.}
  
  \label{fig:fig6}
\end{figure}




\end{document}